\documentclass[letterpaper,twocolumn,11pt]{article}
\usepackage{amsmath,url}
\usepackage{graphicx}
\usepackage{cite}

\newcommand{\tronly}[1]{}
\newcommand{\comment}[1]{}

\begin{document}

\date{}

\title{\Large \bf An Oracle and Observations for the OpenAI Gym / ALE Freeway Environment}

\author{James S. Plank \and Catherine D. Schuman \and Robert M. Patton}

\maketitle

\thispagestyle{empty}
\pagestyle{empty}

\subsection*{Abstract}
The OpenAI Gym project contains hundreds of control problems whose
goal is to provide a testbed for reinforcement learning algorithms.
One such problem is ``Freeway-ram-v0,'' where the observations presented
to the agent are 128 bytes of RAM.  While the goals of the project are
for non-expert AI agents to solve the control problems with general training,
in this work, we seek to learn more about the problem, so that we can better
evaluate solutions.  In particular, we develop on oracle to play the game,
so that we may have baselines for success.  We present details of the oracle,
plus optimal game-playing situations that can be used for training and testing
AI agents.

\section{Introduction}

The OpenAI Gym is an open-source, python based environment whose goal is
to enable research on reinforcement learning.  It focuses on control problems,
where an agent is presented {\em observations}, and must provide {\em actions} to the 
environment.  The result of the actions are rewards, which an agent seeks to maximize.

In this paper, we focus on the environment called ``Freeway-ram-v0.''  This is an
Atari 2600 game, simulated through the ``ALE'' environment~\cite{bnv:13:ale}.  In the game,
the player is represented by a chicken, that must repeatedly cross ten lanes of traffic,
in which cars are traveling horizontally at a variety of speeds and directions.  
The chicken starts at the bottom of the screen, and must get to the top.
Once it gets to the top, it starts again at the bottom.
If the chicken collides with a car, it automatically moves down a lane or two, and is
paralyzed for a brief period of time.

The observations
presented to the agent are composed of a portion of the game's memory -- 128 unlabeled bytes.
There are three actions that are input to the game as 0, 1 and 2:

\begin{itemize}
\item 0: Stay put.
\item 1: Move up.
\item 2: Move down.
\end{itemize}

The player receives a point each time the chicken reaches the top of the screen.  The game
runs for roughly two minutes and fifteen seconds.

This paper presents information that is useful to anyone
who seeks to utilize this environment for machine learning.
Our goal is not to work on a general AI agent.  In fact, it is the opposite --
we seek to understand this application and its optimal behavior. Our goal is to
better understand the limits of AI agents, and to develop non-general agents using
experimental technologies.  In doing so, we hope to understand how to design more general
agents for these technologies that in the end will not need to be informed by expert
information.

Our results from this paper are as follows.  First, we identify various components of the
observations, so that our players may be simplified.  Second, we describe how the application
works empirically.  This allows us to turn game state into a graph, and formulate an
optimal solution to the game as a shortest path problem.  We solve the shortest path
problem with the A-Star algorithm, and present two sets of results.  
The first
is a collection of 10,000 distinct game-playing scenarios, where the objective is to have
the chicken cross the road once, in the shortest amount of time.  
The second is
a collection of 25 optimal complete games, each starting with a different game seed.  
We believe that the first set of 
scenarios present better opportunities for machine learning than the second.  However,
literature that includes this environment uses the second to present their results.

Our data is included in a Github repository, so that others may repeat and/or
leverage our work.  This repository is at \url{https://github.com/jimplank/freeway-oracle}.

\section{The Freeway Application - General Observations}

\begin{figure}[ht]
\begin{center}
\includegraphics{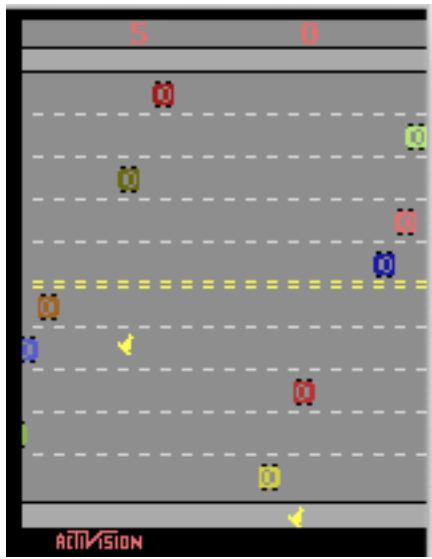}
\caption{\label{fig:screenshot} Screen shot of the Freeway environment in mid-game.}
\end{center}
\end{figure}

We present a screen shot of a game in Figure~\ref{fig:screenshot}.
There are two chickens -- one on the left and one on the right.  The one on the
right does not move.  The one on the left is in the 3rd lane of traffic (zero
indexed from the bottom).  In lanes 0 through 4, the cars move from left to right,
wrapping around, and in lanes 5 through 9, they move from right to left.

To make some inferences about the 128 bytes of observations, we programmed some
simple agents that blindly moved up at each time step, and then we graphed each 
of the 128 bytes over the course of the game.  Those graphs are included in the
Git repository.  From these graphs, we infer the following pieces of information:

\begin{itemize}
\item Byte 14 (zero indexed) contains the Y position of the player.  It starts
      at a minimum value of 6, and can achieve a high value of 177.  Any value
      175 or greater is considered "crossing the road."
\item Bytes 108 to 117 contain the X positions of the cars.  These range from
      0 to 159.  The cars all start at X=0 at the beginning of the game.
\item Cars 0 through 4 move from left to right with average speeds of 0.6,
           0.75, 1.0, 1.5 and 3.0 units per timestep.
\item Cars 5 through 9 move from right to left with average speeds of 3.0,
           1.5, 1.0, 0.75 and 0.6 units per timestep.
\item Although the average speeds of the cars remain constant, the actual X
      positions of the cars change randomly and non-uniformly.  They are,
      however, deterministic according to the game seed.  Specifically,
      the X positions of the cars are precisely determined by the seed and 
      the timestep.   Thus, the movement of the cars looks very similar from
      game to game, although the actual X values change, and that can have a profound
      impact on the player's moves.
\item Byte 106 is a "cooldown".  When its value is zero, then the chicken
      will react to input.  When its value is non-zero, then the chicken will not
      react to input.  Its value increases from zero on a collision with
      a car, and when the chicken's Y value reaches or exceeds 175.  When it
      is greater than zero, it decreases from timestep to timestep.
\item In general, an action of one increases the chicken's Y value, an action 
      of zero keeps it the same, and an action of two decreases it.  The actual
      amount is random, but does not exceed a value of 4, and it averages 3.25.  
      It is deterministic
      according to the seed and the previous sequence of actions.  It is not
      deterministic according to the seed and Y value of the player, or the 
      seed, timestep and Y value of the player (which is unfortunate).
\item The chicken collides with a car when it is in middle of the 
      the car's lane, and the car
      has an X value between 40 and 53.  When the chicken is at the edge's of the
      car's lane, then the X values are more restricted, because the cars collide
      with the chicken's pointy top or bottom.
\item Each car's lane is 17 units in width.  Car $i$'s lane starts at $16i+13$,
      and ends at $16(i+1)+13$.  That means that adjacent lanes overlap by one
      unit.
\item When the chicken collides with the car, it moves down 1 to 2 lanes, and
      is paralyzed for some number of timesteps.  When the chicken gets to the
      top, it eventually resets its Y value to 6, and is paralyzed for some number
      of timesteps.
\item The total number of timesteps in the game varies from seed to seed, and is
      typically between 2700 and 2799.
\end{itemize}

If the player simply repeats an action of one, the score will be between 21 and 24,
depending on the seed.  The distribution of scores for seeds 0 through 999 
is displayed in Table~\ref{tab:one}.  The raw data is in the Git repository.

\begin{table}[ht]
\begin{center}
\begin{tabular}{c|c}
Score & Number of seeds \\
\hline
21 & 779 \\
22 & 0 \\
23 & 204 \\
24 & 17 \\
\end{tabular}
\caption{\label{tab:one} Distribution of scores when the player simply enters an
action of one at every time step.  The data is over 1,000 games with seeds from 
zero to 999.}
\end{center}
\end{table}

In Section~\ref{sec:other}, we note some papers that present results on the Freeway app.

\section{An Oracle for the Environment}

Our goal in writing an oracle is to be able to inform machine learning
researchers on the best possible performance of an agent on the environment.
To do this, we consider a run of the problem to be a graph, and use the
A-Star algorithm to repeatedly find the shortest path from a Y position of
6 to a Y position that is greater than or equal to 175.

We formalize the process below, noting that there is one limiting assumption
that we make.  For a given starting seed, we consider each combination of timestep and Y position to be a node on a graph.  Fortunately, the locations of the cars are deterministic according to the starting seed and the timestep, so we
do not have to keep track of these.  

For each node on the graph, there are three potential outgoing edges -- one
for each possible action.  Each edge has a weight of one, representing one
timestep.  Thus, this is a straightforward graph problem.  
The challenge, however, is constructing the graph.  The reason is that the
new Y value that results from an action is random, and only deterministic
based on the starting seed and the sequence of actions that precede it.  
What this means is that if there are two paths to a single node, then the
edges going out of the node may differ, even if the action from the node is
the same.

\begin{figure*}[ht]
\begin{center}
\includegraphics{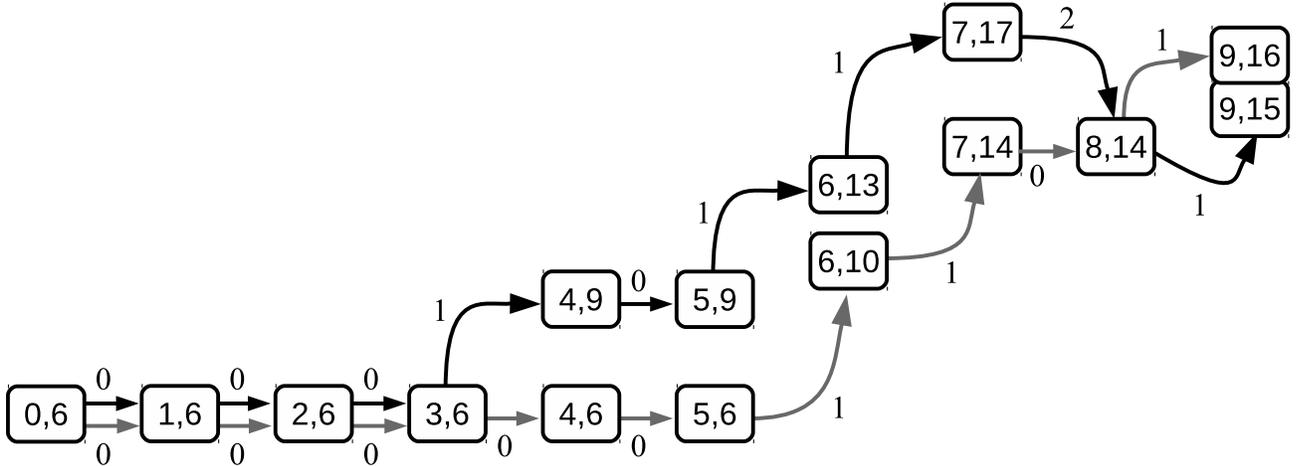}
\caption{\label{fig:exg1} An example of two paths from the initial starting state (0,6) when the
seed is zero.  Nodes are labeled (timestep,Y-value), and edges are labeled with their actions.
Both paths arrive at the state (8,14) and take an action of one (move up).  One path moves up
two Y units to (9,16), while the other path only moves up one, to (9,15).  This fact complicates
the process of finding an optimal path through the graph.}
\end{center}
\end{figure*}

We show an example in Figure~\ref{fig:exg1}.  In this figure, we run the environment twice,
both with a seed of zero, but with a different sequence of actions.  As shown in the graph,
both paths arrive at node (8,14) and take an action of one, to move up.  One path moves up 
two Y units to (8,16), while the other only moves up one.  This demonstrates a challenge
in writing an oracle, in that labeling nodes solely by their timesteps and Y values is
insufficient.  A node should also be labeled with the path that led to it.  However, doing so
would blow up the graph.

Our solution to this problem is to limit the graph so that there is only one node for each 
(timestep, Y-value) pair.  Each of these nodes is annotated with a back edge to the node that
first reached the node in the shortest path calculation.  Thus, there is exactly one stored path to
each node in the graph, and that path is used to arrive at the node whenever it is processed in the
shortest path calculation.  In this way, the edges out of the node are always the same.

This limitation means that we may not actually find the shortest path through the graph, because
we may choose a path to a node (such as the black path to (8,14) in Figure~\ref{fig:exg1}),
whose actions are inferior to a different path (such as the gray one).  However, the differences
in values are slight, and we surmise that while our results may be improved slightly, the improvement
will be minimal.

In our calculation, we disallow collisions.  If a node represents a collision state, then there
are no edges coming out of it.  If a node has a Y value greater than or equal to 175, then it is 
a terminal node.

We use the A-Star algorithm to perform the shortest path calculation.  A-Star requires an estimated
distance from each node to a terminal node, and in order for the algorithm to be correct, the
estimated distance cannot be larger than the actual distance.  In our calculation, our estimate
assumes that the path from a node to a terminal node will travel four Y units on each time step
and never have a collision.  Since the maximum amount one moves up is four units, this estimate is
sufficient for the calculation, and it does significantly reduce the size of the graph from, for
example a breadth-first search that is the non-heuristic solution to the shortest path problem.

The program is partitioned into two pieces.  The first piece is written in C++, and drives the A-Star
calculation, storing the graph in memory as it is created.  The second piece is written in python,
using the OpenAI Gym environment to drive the Freeway program.  Whenever the first program needs
to process a node in the A-Star algorithm, it first checks its stored graph to see if the information
is present.  If it is, then it may proceed without contacting the second program.  If the information
is not present, then the first program communicates with the second program over a Unix pipe, presenting 
the entire path to the node in question, so that the python program may rebuild the path to the node,
resulting in consistent actions from the node.

This is a time-consuming process, as the python program takes on the order of seconds to restart
the environment and reconstruct a path.  To help reduce the number of times that the python
has to reconstruct paths, when a node is processed, the python program continues processing a 
path from that node composed solely of ``up'' actions, until it reaches a collision or a terminal
node.  In that way, it needs to be called fewer times.

During the A-Star calculation, edges are processed in the order ``up'', ``stay'', ``down.''  The
reason is to favor paths that have ``up'' actions, as consecutive ``up'' actions seem to result
in more distance covered than mixing the actions.

Each calculation for Section~\ref{sec:single} took between an hour and 15 hours to run on 
the TENNLab cluster (2.5 GHz AMD cores).
For Section~\ref{sec:all}, they took between 5 and 6 days.  We could
have sped them up significantly by running multiple python processes and farming out the 
work; however, as we were solely interested in the results presented herein, we opted to 
keep the program simple and endure the large running times.

\begin{figure}[ht]
\begin{center}
\includegraphics[width=3in]{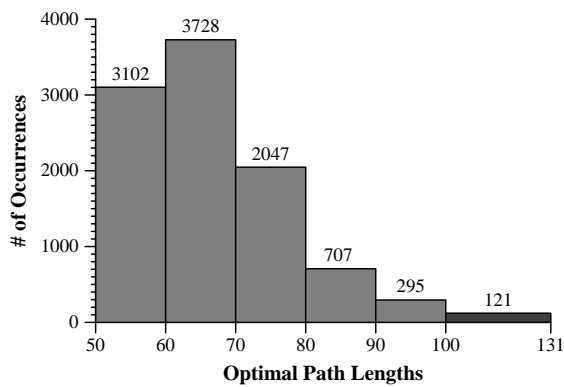}
\caption{\label{fig:histo} Histogram of optimal path lengths in Experiment \#1,
which shows the fastest time for the chicken to cross the freeway at
different starting seeds and starting timesteps.}
\end{center}
\end{figure}

\section{Experiment 1: How fast can the chicken cross the road?}
\label{sec:single}

For our first experiment, we performed 10,000 runs, where each run is parameterized by a 
seed from 0 to 999,999 and a starting timestep from 0 to 2,500.  The Freeway application is
then given this seed, and the appropriate number of ``stay'' actions to get to the given 
timestep.  At that point, the A-Star calculation is executed to determine the minimum number
of timesteps for the chicken to cross the road once.

While this is not the canonical use of the Freeway application, it may be more appropriate for
machine learning algorithms.  The reason is that these individual runs have more variety
than simply running the application from start to finish.  The required timesteps have a great
deal more variety, and may do a better job of isolating behaviors.

To characterize these runs, in Figure~\ref{fig:histo}, we show a histogram of
the optimal path lengths for the 10,000 runs.  The shortest path length is 50
(exactly one run), and the longest path length is 131 (also exactly one run).
Of the 10,000 runs, 2,326 have optimal paths composed solely of ``up'' actions. 
That means that 76.7 percent of the runs require the chicken to take some action
that is either ``stay'' or ``down.'' 

\begin{figure}[ht]
\begin{center}
\includegraphics[width=3in]{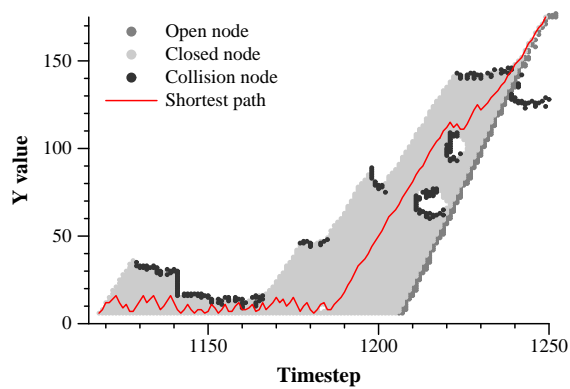}
\caption{\label{fig:166422} Details on the A-Star calculation of the shortest path
when the seed is 166,422 and the starting timestep is 1,118.  The optimum path
is 131 timesteps, and will present a challenge to machine learning systems.}
\end{center}
\end{figure}

The average path length over the 10,000 runs is 66.69.  For reference, simply entering "up"
at each time step yields an average path length of 113.5.

To give some flavor of what the A-Star calculation looks like, in Figure~\ref{fig:166422},
we graph the calculation for the worst case of our data, which is seed 166,422 and
starting timestep 1118, whose optimum path is 131 timesteps.  Besides the shortest path, the
graph displays ``closed'' nodes, which have been processed by the A-Star algorithm, ``open''
nodes which are on the ordered set of nodes to be processed, and ``collision'' nodes, which
must be avoided.  The graph shows that the chicken goes through four phases in its 
crossing of the road:

\begin{itemize}
\item Phase 1 (Roughly time steps 1118 to 1185): Delaying at the beginning while waiting for cars above to not be dangerous.
\item Phase 2 (Time steps 1185 to 1220): Crossing the road with haste, to get above the 
car at Y value 110.
\item Phase 3 (Time steps 1220 to 1230): Delaying for the car at Y value 140 to pass.
\item Phase 4 (Time steps 1230 to 1248): Crossing again with haste, to get between the
cards at Y values 130 and 140.
\end{itemize}

The graph also shows the effectiveness of A-Star, which eliminates the processing of 
a large number of nodes to the right of and below the line of ``open'' nodes.  These nodes
would have to be considered by a breadth-first search or Dijkstra's shortest path 
algorithm.

The 10,000 paths are included in the Github repository.  We believe that the data
set will be useful for machine learning applications that can partition the 10,000
runs into training and testing sets.

\section{Experiment 2: What's the best high score?}
\label{sec:all}

For this determination, we started at timestep zero, and
we used A-Star to determine the fastest path to a Y value greater than or equal to 175.
Then we deleted the graph and continued to perform "up" actions, until the chicken was back
in a starting Y value of 6.  We then commenced the A-star algorithm again.  Note that for
determinism, we always use the same path to a node in the graph, 
which includes previous crossings of the road.  Thus, this experiment takes quite a bit of time,
because of the replaying of the paths.  As mentioned above, 
on our cluster, each run took between 5 and 6 days
to complete.

We performed 25 runs, with starting seeds of 0, 1, 2, 3, 4, and every multiple of 50 up to 1000.
Each game had a final optimal score of 34.  
We include all of the paths for these runs in \url{https://github.com/jimplank/freeway-oracle}.
We also include a movie containing all 25 runs at~\url{https://www.youtube.com/watch?v=FKYXuzNVZKg}.

\begin{figure*}[ht]
\begin{center}
\includegraphics[width=6in]{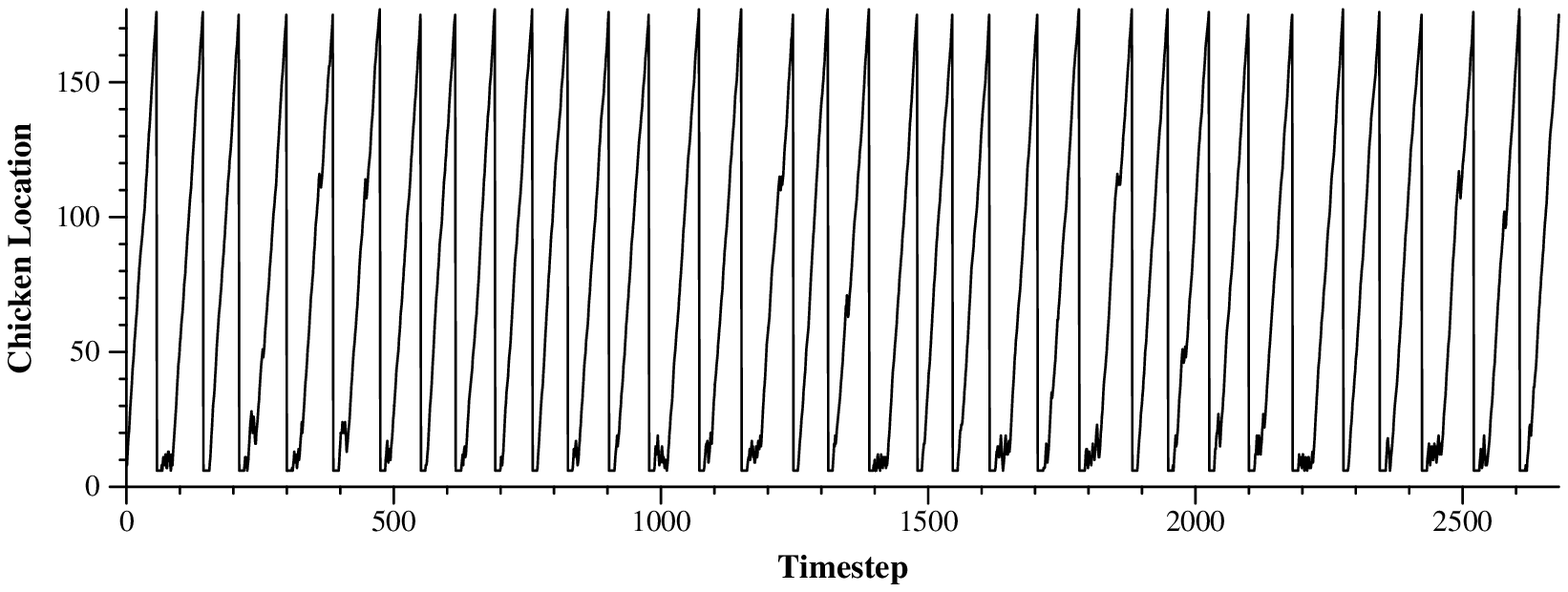}
\caption{\label{fig:e2} The Y values of the chicken in an optimal playing
of the environment for seed zero.}
\end{center}
\end{figure*}

In Figure~\ref{fig:e2}, we show the 34 crossings for seed zero.

\section{Other Work}
\label{sec:other}

Bellemare {\em et al} show a best agent score of 19.1, using the RAM.  Using the pixels,
it's 16.4.  Their "const" agent scores 21 (which means that they didn't test multiple
seeds).  Their "perturb" agent scores 22.5. \cite{bnv:13:ale}

Machado {\em et al} reports between 31.8 and 32.1 for "Sarsa + Blob-Prost" and 33.0 for DQN.  
It also says that there are different modes that can be set in the new version of ALE, and
that freeway has 8 of these.~\cite{mbt:18:ral}

DeepMind reports scores of 29.1 and up to 34 on the Freeway application \cite{hmv:18:rci}.

\section{Spiking Neural Networks with Tennlab}

Within the TENNLab framework~\cite{psb:18:ten}, we were able to get scores between 32
and 33 by employing spiking neural networks trained with both genetic and Monte Carlo searches.  
In these runs, we preprocess the
data so that the observations of the cars are relative to the chicken.  We have not published
these findings yet.  

One of the neural networks has just five neurons: two inputs (the two closest cars),
three outputs and just two edges.  The ``Up'' output neuron has no incoming edges,
but since ties among the outputs are settled in favor of ``Up'', when no outputs spike, the
action is ``Up''.  This network scored over 32.

As a final remark, one interesting network that we generated in this search, we term
``the scared chicken.''  Its score is poor: 16.9.  However, its behavior is interesting
as it is fraught with caution, nicely evading cars, while not crossing the street very
well!  We have posted a video of this network at \url{https://www.youtube.com/watch?v=B_nQtRMYVIc}.

\section{Acknowledgements}

This research was funded by a grant from UT-Battelle.

\bibliographystyle{plain}
\bibliography{references}

\end{document}